\documentclass[dvipsnames,format=sigconf,anonymous=false,review=false]{acmart}
\usepackage{multirow}
\usepackage{multicol}
\usepackage{graphicx}
\usepackage{subcaption}
\usepackage{amsmath}
\usepackage{algorithm}
\usepackage{algpseudocode}
\usepackage{amsfonts}
\usepackage{orcidlink}
\usepackage{xcolor}
\AtBeginDocument{%
  }

\copyrightyear{2025} 
\acmYear{2025} 
\setcopyright{none} 
\acmConference[GECCO '25 Companion]{Genetic and Evolutionary Computation Conference}{July 14--18, 2025}{Malaga, Spain} 
\acmBooktitle{Genetic and Evolutionary Computation Conference (GECCO '25 Companion), July 14--18, 2025, Malaga, Spain}\acmDOI{10.1145/3712255.3734302} 
\acmISBN{979-8-4007-1464-1/2025/07}

\begin{document}

\title{A Better Multi-Objective GP-GOMEA - But do we Need it?}

\author{Joe Harrison \orcidlink{https://orcid.org/0000-0002-3427-5251}}
\email{Joe.Harrison@cwi.nl}
\affiliation{%
  \institution{Centrum Wiskunde \& Informatica}
  \city{Amsterdam}
  \country{The Netherlands}
}

\author{Tanja Alderliesten \orcidlink{https://orcid.org/0000-0003-4261-7511}}
\affiliation{%
 \institution{Leiden University Medical Center}
 \city{Leiden}
 \country{The Netherlands}}

\author{Peter A.N. Bosman \orcidlink{https://orcid.org/0000-0002-4186-6666}}
\affiliation{%
  \institution{Centrum Wiskunde \& Informatica}
  \city{Amsterdam}
  \country{The Netherlands}
}
\affiliation{%
  \institution{Delft University of Technology}
  \city{Delft}
  \country{The Netherlands}
}

\renewcommand{\shortauthors}{Harrison et al.}

\begin{abstract}
In Symbolic Regression (SR), achieving a proper balance between accuracy and interpretability remains a key challenge. The Genetic Programming variant of the Gene-pool Optimal Mixing Evolutionary Algorithm (GP-GOMEA) is of particular interest as it achieves state-of-the-art performance using a template that limits the size of expressions. A recently introduced expansion, modular GP-GOMEA, is capable of decomposing expressions using multiple subexpressions, further increasing chances of interpretability. However, modular GP-GOMEA may create larger expressions, increasing the need to balance size and accuracy. A multi-objective variant of GP-GOMEA exists, which can be used, for instance, to optimize for size and accuracy simultaneously, discovering their trade-off. However, even with enhancements that we propose in this paper to improve the performance of multi-objective modular GP-GOMEA, when optimizing for size and accuracy, the single-objective version in which a multi-objective archive is used only for logging, still consistently finds a better average hypervolume. We consequently analyze when a single-objective approach should be preferred. Additionally, we explore an objective that stimulates re-use in multi-objective modular GP-GOMEA.
\end{abstract}




\maketitle

\section{Introduction}
Symbolic Regression (SR) is an important eXplainable Artificial Intelligence (XAI) technique, where the goal is to uncover the underlying relationships between input variables and output targets in a given dataset through the construction of symbolic expressions. Symbolic expressions have the potential to be interpretable. As machine learning becomes more integrated into decision-making processes in critical domains such as healthcare, finance, and criminal justice, the need for interpretable models grows, making SR algorithms an essential tool in the XAI landscape~\cite{van2022explainable, doshi2017towards, lipton2018mythos}.

In XAI, the accuracy of machine learning models is not the only important objective; interpretability is equally important. The interpretability of models is influenced by various factors, including the size of the expression, the number of consecutive compositions, the types of operators employed, and the capabilities of the person interpreting the expression~\cite{virgolin2020learning,kommenda2015complexity}. Unlike traditional regression methods that primarily focus on parameter estimation, SR simultaneously searches for both the structure and parameters of symbolic expressions~\cite{kommenda2020parameter}. Consequently, SR naturally lends itself to Multi-Objective (MO) optimization, where solutions with trade-offs between competing objectives, such as accuracy and expression complexity, can be explicitly searched for. 


Recently, it was found that using classic selection based on non-dominated sorting and subsequent variation to generate a new offspring population, as e.g., in NSGA-II, is prone to evolvability degeneration~\cite{liu2022evolvability}. The population then tends to quickly get flooded with many small solutions, since this objective is easy to optimize, making it more difficult to subsequently find larger expressions.

GP-GOMEA, a state-of-the-art population-based approach to GP~\cite{virgolin2021improving,schlender2024improving}, differs from classic selectorecombinative EAs in that it employs a technique called optimal mixing, which is more akin to a local search approach. Moreover, the first MO version of GP-GOMEA~\cite{sijben2022multi} follows the MO-GOMEA structure in which clustering is used to spread the search bias across the approximated Pareto front. Among the clusters, extreme clusters are identified in which improvements for only one objective are accepted. Optimal mixing combined with clustering essentially mitigates evolvability degeneration. However, different versions of GOMEA exist, for different types of optimization problems, in which clustering is approached differently, e.g.,~\cite{luong2014multi,bouter_multi-objective_2017}. Particularly, they include a different way to balance the clusters, which could be important when objectives have varying difficulty. We therefore want to study if alternative clustering approaches can have added value in GP-GOMEA.

It is well known that GP tends to bloat, as larger expressions often achieve higher accuracy~\cite{o2009riccardo}. Consequently, a single-objective (SO) optimization process, which solely focuses on accuracy, may naturally favor increasingly larger expressions over time, especially when starting with relatively small ones. This could potentially result in a non-dominated front with a higher average hypervolume compared to MO GP-GOMEA, as the latter balances both accuracy and expression size. In this paper, we investigate whether and to what extent this effect occurs in GP-GOMEA.


Recently, a modular version of GP-GOMEA was introduced in~\cite{placeholdermodularpaper} that allows for the efficient evolution of larger expressions with GP-GOMEA by virtue of re-using subexpressions as functions or input features. Re-use of subexpressions can at the same time potentially lower the complexity of expressions because they can be interpreted in a faceted manner, understanding the subexpressions separately and within the whole, without losing much in accuracy. However, a challenge remains in that limited function re-use was observed in the original version of Modular GP-GOMEA, which is single-objective~\cite{placeholdermodularpaper}. We hypothesize that the implementation of a special type of parsimony pressure that discounts subexpression re-use in multi-objective optimization may positively influence this aspect, potentially leading to more re-use of expressions.

In this paper, we propose improvements to the MO clustering method for MO Modular GP-GOMEA. We demonstrate that the accuracy-complexity trade-off presents a scenario where SO optimization should be employed. We give insight into when SO or MO Modular GP-GOMEA should be used in light of other objectives in their trade-off with accuracy. Furthermore, we investigate the effects of parsimony pressure on the frequency of functional re-use within the generated expressions.

\section{Methods}
This section begins with an explanation of the general functioning of GP-GOMEA, followed by an explanation of Modular GP-GOMEA, and concludes with a detailed description of how these components are integrated into MO Modular GP-GOMEA.
\subsection{GP-GOMEA}
\label{sec:GPGOMEA}
\begin{figure}[h]
  \vspace*{-4mm}
  \centering
  \includegraphics[width=\linewidth]{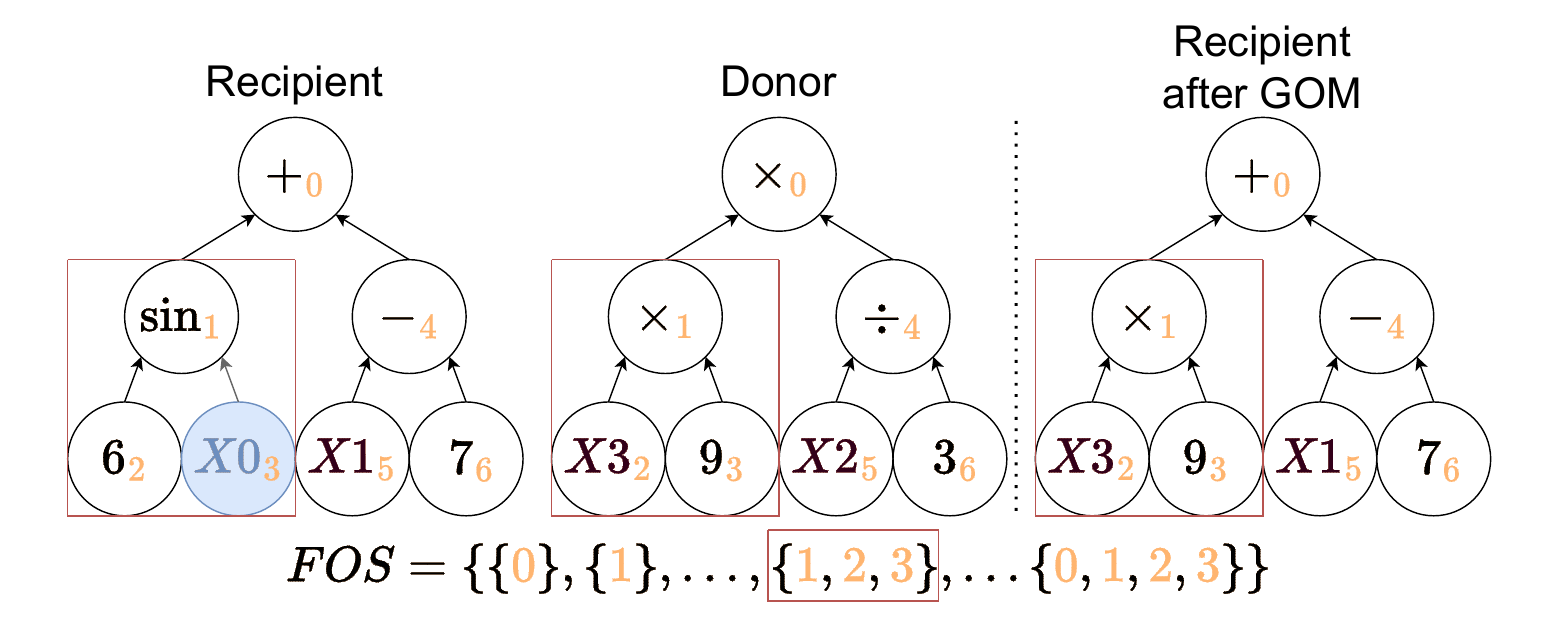}
  \vspace*{-7mm}
  \caption{Example of the GOM procedure. Orange subscript numbers denote string indices from the tree's pre-order traversal. Numbers in subsets correspond to tree indices. Tree sections of the recipient and donor matching the subset in the FOS are outlined in red, with introns shaded in blue.}
  \label{fig:GOMprocedure}
  \vspace*{-2mm}
\end{figure}

Symbolic expressions in GP are often modelled as binary-unary trees. Unlike traditional GP~\cite{koza1992genetic}, which allows trees of varying sizes, GP-GOMEA uses a fixed tree template. This allows individuals in GP-GOMEA to be represented as strings, where each string index bijectively maps to a position in the tree. The mapping is determined using the pre-order traversal of the tree template (see Figure \ref{fig:GOMprocedure}). 

In GP-GOMEA, linkage information is attempted to be leveraged via the identification of dependencies between tree positions. As a surrogate measure of dependence, Mutual Information (MI) between all pairs of positions as measured in the population, is used. The Unweighted Pair Group Method with Arithmetic Mean (UPGMA)~\cite{gronau2007optimal} is then applied to the MI matrix to hierarchically cluster positions, constructing a Family of Subsets (FOS) in the shape of a tree commonly called the linkage tree. For further details, see~\cite{virgolin2021improving}.

In GP-GOMEA, variation is achieved through Gene-pool Optimal Mixing (GOM). GOM is applied to every individual in each generation. In GOM, a clone of the individual is created, and for each subset in the FOS, the corresponding genes in the clone are replaced with those of a donor randomly chosen from the population (see Figure~\ref{fig:GOMprocedure}). If the change swaps meaningful, non-intron, genes, and does not lead to worse fitness, it is accepted. After processing all FOS subsets, the resulting individual is added to the offspring set. Once all individuals are processed, the offspring set replaces the population. As in~\cite{virgolin2022coefficient}, coefficients are also mutated after GOM, with any mutation being reverted if it worsens fitness.

\subsection{Modular GP-GOMEA}
\begin{figure}[h]
  \vspace*{-4mm}
  \centering
  \includegraphics[width=\linewidth]{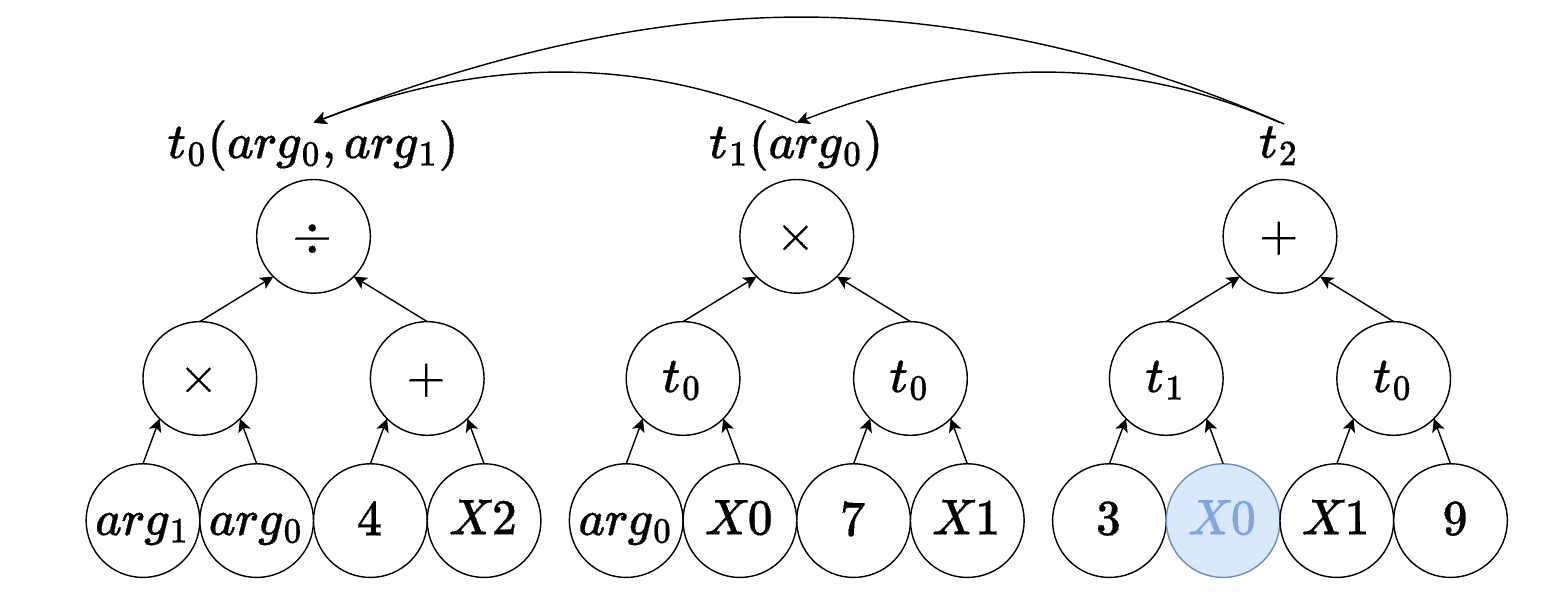}
  \vspace*{-7mm}
  \caption{Example of an individual in Modular GP-GOMEA and how trees in its multi-tree representation can call each other. The last tree $t_2$ is the top-level expression and can call preceding trees $t_1$ and $t_0$, but not vice versa. Nodes $X_1$ and $9$ in $t_2$ are the input arguments $arg_0$ and $arg_1$ in $t_0$ respectively.}
  \label{fig:modularexample}
  \vspace*{-2mm}
\end{figure}

In this paper, we use Modular GP-GOMEA~\cite{placeholdermodularpaper}, a recently introduced multi-tree variant of GP-GOMEA where trees can call other trees as (functional) subexpressions (see Fig.~\ref{fig:modularexample}). This allows for improving accuracy without affecting efficiency compared to GP-GOMEA with a larger single template. To avoid cyclic calling of subexpressions, only trees $i$ that precede a tree $j$ in the multi-tree can be called. The left and right inputs of subexpression nodes are the input arguments $arg_0$ and $arg_1$, respectively, in the subexpression tree.

For each tree in the multi-tree, a separate FOS is constructed (e.g., for the example in Figure~\ref{fig:modularexample}, there is a separate FOS for each of $t_0, t_1$, and $t_2$). For each tree, except the last (top-level expression) tree, the FOS for that tree contains a subset with all indices in the tree so as to allow full subtrees to be swapped. Before applying GOM, all subtree FOSes are combined into one large FOS and given a subscript to keep track to which tree in the multi-tree each subset pertains, e.g., $FOS=\{\{0,1,2\}_0, \{4,5\}_1, ...\}$. Upon applying GOM to an individual, the combined FOS is shuffled and a GOM step is performed for each FOS element in turn.

\subsection{Multi-Objective GP-GOMEA}
\label{sec:MOGPGOMEA}
\begin{figure}[h]
  \vspace*{-5mm}
  \centering
  \includegraphics[width=0.63\linewidth]{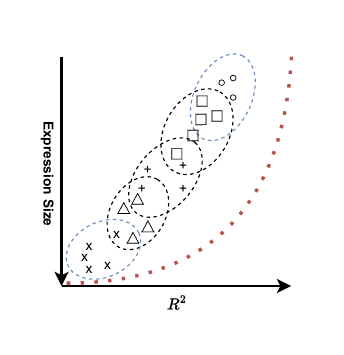}
  \vspace*{-10mm}
  \caption{Example of clusters (similar symbols) and corresponding donor clusters (dashed ovals). $\times$ and $\circ$ are extreme clusters for the objectives expression size and $R^2$ respectively. The blue dashed ovals are the corresponding donor clusters. The red dotted line represents the Pareto front.}
  \label{fig:donorpop}
  \vspace*{-4mm}
\end{figure}
 

A key driver to promote spreading the search bias of an EA along the Pareto approximation front, is to use clustering. The first introduced version of MO-GP-GOMEA~\cite{sijben2022multi} follows the first multi-objective version of optimization-based GOMEA~\cite{luong2014multi}. The clustering approach used there, was based on the Balanced k-Leader-Means algorithm (BKLM) introduced in~\cite{bosman2010anticipated}. Two types of clusters are identified: (1) extreme clusters for which Single-Objective (SO) GOM is performed based on their corresponding objective, and (2) middle clusters for which MO GOM is performed. In MO GOM, changes are accepted in case a Pareto improvement is found, when all objectives stay the same, or when a change leads to a solution that is accepted into the elitist archive~\cite{luong2012elitist}. In this paper, we use $k=5$ clusters. Because we furthermore consider two objectives here, we have two extreme and three middle clusters, see Figure~\ref{fig:donorpop} for an example.

To create the clusters, first, the solution with the best objective value for a randomly chosen objective is chosen. Then, the leaders of the remaining $k-1$ clusters are chosen from the remaining population by iteratively picking the individual with the largest Euclidean distance in normalized objective space to the other leaders~\cite{luong2014multi}. Then, while there remain unpicked individuals, the clusters are iterated over in random order and the individual that is closest to the cluster leader is assigned to that cluster. Once every individual is assigned to a cluster, $k$-means clustering is performed. Afterward, the cluster centers are used to first construct donor clusters by choosing the $\frac{2n}{k}$ individuals closest to each cluster center, where $n$ is the population size. These clusters all have equal size to ensure control over the numbers of individuals from which linkage is learned. Indeed, for each donor cluster, a separate FOS is learned for each tree in the multi-tree representation (i.e. $5\times4=20$ FOSes in the case of this paper). In GOMEA, each individual must undergo GOM, but the clustering procedure thus far may have left individuals unassigned or assigned to multiple clusters. Individuals without an assigned cluster are assigned to a random cluster. For individuals with multiple assigned clusters, ties are broken randomly. The clusters for which their center represents the best objective value for one of the objectives, becomes the extreme cluster for that respective objective.

\begin{algorithm}
\caption{\texttt{Balanced k-(2-)leader-means}}
\label{alg:balancedkleadermeans}
\begin{algorithmic}[1]
\State Normalise objective values for all individuals.
\State \textcolor{blue}{\textbf{for} each objective in random order \textbf{do}}
    \State \quad \textcolor{blue}{Sort population in objective.}
    \State \quad \textcolor{blue}{Assign top $\frac{n}{k}$ individuals to an extreme (SO) cluster.}
    \State \quad \textcolor{blue}{Remove these individuals from the pool of individuals.}
\State \textcolor{blue}{\textbf{end for}}
\State \textcolor{blue}{Set $k$ to $k-2$.}
\State Choose a random objective $o$.
\State Initialize first leader as individual with best value for $o$.
\While{fewer than $k$ leaders selected}
    \State Compute distances from remaining candidates to leaders.
    \State Select candidate with maximum distance as new leader.
\EndWhile
\State Assign individuals to nearest leader to form initial clusters.
\State Perform k-means clustering.
\State Undo all cluster assignments, but keep cluster centers.
\State \textcolor{purple}{\textbf{while} Individuals not assigned to a cluster \textbf{do}}
    \State \quad \begin{minipage}[t]{75mm}\textcolor{purple}{Loop over $k$ clusters in random order and assign to each cluster the unassigned individual that is closest to the cluster center, without updating the cluster center.}\end{minipage}
\State \textcolor{purple}{\textbf{end while}}
\State \textcolor{olive}{For each cluster center, create donor cluster with closest $\frac{2n}{k}$ individuals.}
\State \textcolor{olive}{Identify clusters as extreme or middle clusters}.
\State \textcolor{purple}{Identify clusters as extreme or middle clusters.}
\State \textcolor{olive}{Create final clusters from donor clusters by assigning unassigned individuals to random clusters and breaking ties for individuals with multiple cluster assignments randomly.}
\end{algorithmic}
\caption{Combined description of various clustering methods in MO (Modular) GP-GOMEA. The original clustering method is defined by all steps in black and \textcolor{olive}{olive}. The BKRR clustering method introduced in this paper is defined by all steps in black and \textcolor{purple}{purple}. The BKmRR clustering method introduced in this paper is defined by all steps in black, \textcolor{blue}{blue}, and \textcolor{purple}{purple}.}
\end{algorithm}

We take inspiration from~\cite{luong2012elitist} and use a method similar to Adaptive Grid Discretisation to keep the number of individuals in the elitist archive at a manageable size. The objective space is divided into grid cells. Individuals are only admitted to the elitist archive if they are non-dominated and occupy an empty grid cell or if they dominate the individual currently residing in the same grid cell. We evenly space the grid with 100 steps between the minimum and maximum value of the elitist archive of the previous generation. We opted for an adaptive method because it is unknown beforehand how large an individual will grow and size is one of our objectives.

\subsection{MO GP-GOMEA vs SO GP-GOMEA with an MO archive}
We distinguish between MO and SO GP-GOMEA (respectively referred to as MO and SO in tables). While the sole objective in SO GP-GOMEA is $R^2$ (maximization), we also maintain an MO elitist archive. This MO elitist archive only \emph{tracks} non-dominated solutions when considering size besides $R^2$, i.e., \emph{the MO archive does not influence the evolutionary process in SO optimization in any way}.

\section{Experiments and Results}
We conduct experiments to address six research questions. All are studied in the context of the modular version of GP-GOMEA, even though we omit the term modular in most of the remainder of this paper. The experiments for each question may build on the findings of prior ones. Consequently, we present the results immediately following the explanation of the setup for each research question:
\begin{enumerate}
    \item Is clustering necessary for MO GP-GOMEA?
    \item How does MO GP-GOMEA compare to SO GP-GOMEA?
    \item What improvements can be derived for MO GP-GOMEA from SO GP-GOMEA?
    \item Does the removal of duplicate solutions affect performance?
    \item Can functional re-use be promoted in MO GP-GOMEA?
    \item How does MO GP-GOMEA perform with other objectives?
\end{enumerate}

\subsection{General Setup}
We run experiments on 5 real-world datasets (see Table~\ref{tab:realworld}). Each experiment is run on a separate core of an AMD EPYC ROME 7282 and is terminated, either when a time budget of 3 hours is reached, or when the MO elitist archive (which is maintained in all experiments) does not change for 100 consecutive generations. The average hypervolume~\cite{zitzler1999multiobjective} over all repetitions is used as a measure of comparison. To allow for easy comparison between datasets we use the coefficient of determination $R^2 = 1-\frac{var(x)}{MSE(x)}$ as a measure of accuracy. For the size of the expression, we expand all subexpression nodes and sum up its non-intron nodes. The hypervolume is computed by normalising the objectives to a range of 0 to 1, achieved by subtracting the minimum objective value and dividing by the maximum objective value after the subtraction step, observed across all experiments, with the exception of $R^2$ which is set to have a fixed minimum and maximum of 0 and 1 respectively. The reference point is set to 0 and 1 for the $R^2$ and size objective respectively. Statistical testing is done using the Wilcoxon signed rank test with Bonferroni-Holm correction (significant results indicated in bold).

\begin{table}[h]
\vspace*{-1mm}
\small 
\tabcolsep=1.5mm 
\centering
\scalebox{0.95}{
\begin{tabular}{lcccc}
\toprule
Dataset & Samples & Features & Mean & Variance \\
\midrule
Airfoil       & 1503   & 5  & 124.8   & 6.9    \\
Bike Daily    & 731    & 11 & 4504.3  & 1935.9 \\
Concrete      & 1030   & 8  & 35.8    & 16.7   \\
Dow Chemical  & 1066   & 57 & 3.0     & 0.1    \\
Tower         & 4999   & 25 & 342.1   & 87.8   \\
\bottomrule
\end{tabular}
}
\caption{Real-world datasets used in our experiments. Two columns that are a linear combination of the target were removed from the Bike Daily dataset.}
\label{tab:realworld}
\vspace*{-5mm}
\end{table}

We initialize the population using the half-and-half method (50\% grow, 50\% full). In the grow method, terminal nodes (input features or coefficients) are sampled with a 50\% probability. Moreover, the probability of sampling a coefficient is then also 50\%. Unused string indices in the fixed-sized tree templates are filled with introns sampled randomly from the operator set. For unary operators the leftmost child node is used (see also Figure~\ref{fig:GOMprocedure}).

Linear scaling (LS) terms are recalculated at every generation for each expression to ensure optimal scaling during the evolutionary process, following the approach proposed by ~\cite{keijzer2003improving}.

In Table~\ref{tab:general_experiment_info} we provide an overview of the general settings that we used in all of the experiments in this paper.

\begin{table}[htbp]
\centering
\resizebox{\columnwidth}{!}{ 
\begin{tabular}{ll}
\toprule
\textbf{Parameter} & \textbf{Setting} \\
\midrule
Population size                  & 4096                     \\
Tree height                      & 4                        \\
\# Multi-trees                   & 4                        \\
Coefficient sampling             & $\sim U(\text{min}_{\text{target}}, \text{max}_{\text{target}})$ \\
Probability sampling coefficient & 50\%                     \\
Function set                     & $+,-,*,/, \sin, \cos, \log, \sqrt,\ \text{subexpression}, \text{arg}_0, \text{arg}_1$ \\
\# Repetitions                   & 10                       \\
Stopping criterion               & 3 hours                   \\
\# Clusters                      & 5                        \\
\bottomrule
\end{tabular}
}
\caption{General experiment settings used in every experiment (unless indicated otherwise).}
\label{tab:general_experiment_info}
\vspace*{-5mm}
\end{table}
\subsection{Is clustering necessary for MO GP-GOMEA?}
\begin{figure}[h]
  \vspace*{-2mm}
  \centering
  \includegraphics[width=\linewidth]{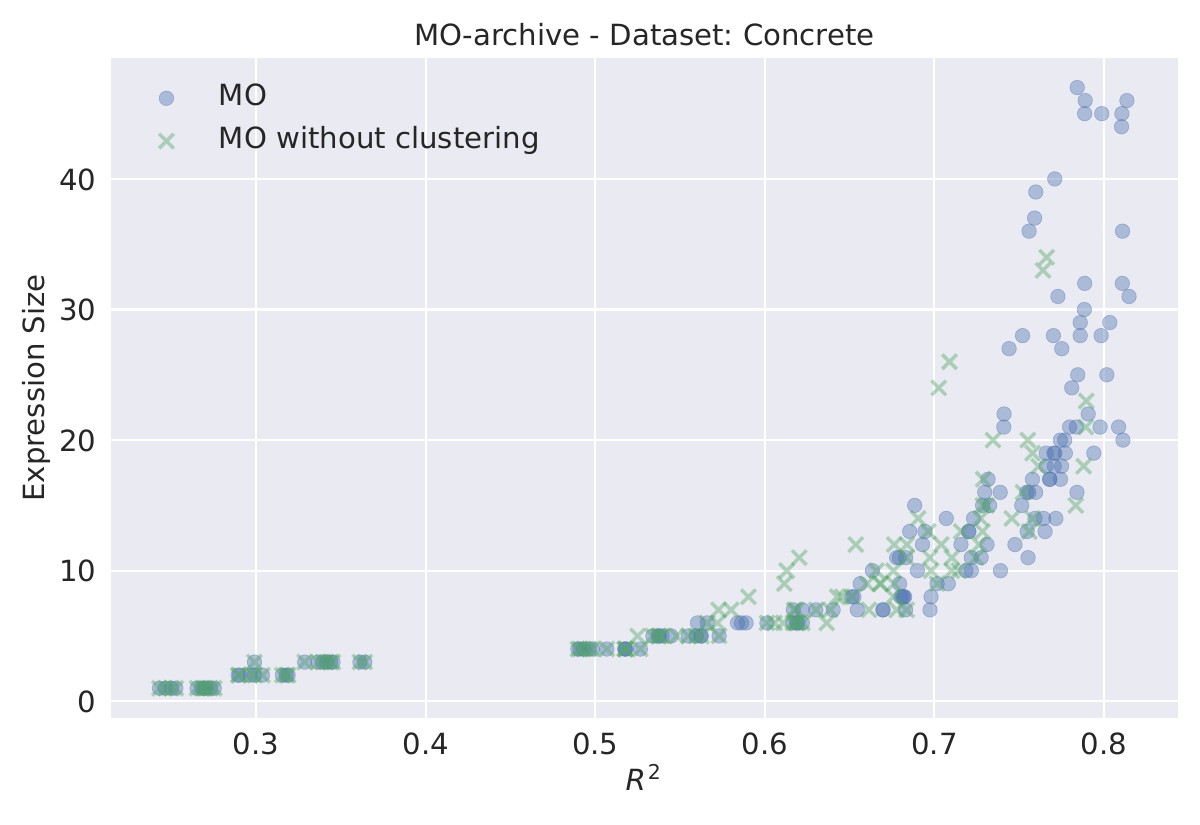}
  \vspace*{-10mm}
  \caption{Individuals in the MO elitist archives in all 10 experiment repetitions.}
  \label{fig:MOcollapse}
\vspace*{-3mm}
\end{figure}

In~\cite{liu2022evolvability} it was shown that with expression size as secondary objective, a classic selection and variation approach to MO GP such as NSGA-II~\cite{deb2002fast} is prone to evolvability degeneration because generating smaller expressions is a much easier objective than improving the accuracy. However, given the more local search nature of GOMEA where small changes are made and immediately tested for improvement, it is not directly clear whether the additional layer of complexity that comes with clustering is required. We therefore first perform an experiment where we compare MO Modular GP-GOMEA using the clustering approach of the first MO GP-GOMEA~\cite{sijben2022multi} to not using clustering.

\begin{table}[h!]
\vspace*{-1mm}
\centering
\scalebox{0.95}{
\begin{tabular}{lcc}
\toprule
\multirow{2}{*}{Dataset} & \multirow{2}{*}{\begin{tabular}[c]{@{}c@{}}MO with\\ clustering\end{tabular}} & \multirow{2}{*}{\begin{tabular}[c]{@{}c@{}}MO without\\ clustering\end{tabular}} \\
\\
\midrule
Air          & 0.696 $\pm$ 0.046 & 0.668 $\pm$ 0.056 \\
Bike         & \textbf{0.839 $\pm$ 0.017} & 0.809 $\pm$ 0.019 \\
Concrete     & \textbf{0.794 $\pm$ 0.011} & 0.741 $\pm$ 0.032 \\
Dowchemical  & \textbf{0.709 $\pm$ 0.036} & 0.636 $\pm$ 0.056 \\
Tower        & \textbf{0.827 $\pm$ 0.021} & 0.746 $\pm$ 0.045 \\
\bottomrule
\end{tabular}
}
\caption{Comparison of average hypervolume ($\pm$ standard deviation) obtained with MO Modular GP-GOMEA with and without clustering. Significant results are indicated in bold.}
\label{tab:nocluster}
\vspace*{-5mm}
\end{table}

The results in Table~\ref{tab:nocluster} show that without clustering the average hypervolume is significantly smaller in 4/5 datasets. In Figure~\ref{fig:MOcollapse} we see that using clustering, especially larger and more accurate solutions can be found within the time budget. We therefore conclude that, while GOMEA does not suffer from evolvability degeneration as does NSGA-II, adding clustering to its procedure is beneficial.

\subsection{How does MO GP-GOMEA compare to SO GP-GOMEA?}
Clustering and performing linkage learning for each cluster separately adds complexity in MO GP-GOMEA. A natural question is to ask whether the added complexity leads to improvements. To this end, we now consider the single-objective version of GP-GOMEA using as sole objective maximizing accuracy ($R^2$).

\begin{table}[h]
\vspace*{-1mm}
\centering
\tabcolsep=2mm
\scalebox{0.95}{
\begin{tabular}{lcc}
\toprule
Dataset           & MO with clustering   & SO            \\
           \midrule
Air          & 0.696 $\pm$ 0.046 & \textbf{0.820 $\pm$ 0.016} \\
Bike         & 0.839 $\pm$ 0.017 & \textbf{0.893 $\pm$ 0.007} \\
Concrete     & 0.794 $\pm$ 0.011 & \textbf{0.855 $\pm$ 0.005} \\
Dowchemical  & 0.709 $\pm$ 0.036 & \textbf{0.853 $\pm$ 0.010} \\
Tower        & 0.827 $\pm$ 0.021 & \textbf{0.899 $\pm$ 0.004} \\
\bottomrule
\end{tabular}
}
\caption{Comparison of average hypervolume ($\pm$ standard deviation) between MO optimization and SO optimization. Significant results are indicated in bold typeface.}
\label{tab:MOSO}
\vspace*{-3mm}
\end{table}

In Table~\ref{tab:MOSO} we see that SO GP-GOMEA significantly outperforms MO GP-GOMEA on all datasets within the limits of our experiments. A larger average hypervolume is achieved with SO GP-GOMEA due to it being able to divert all its search efforts towards finding expressions with high $R^2$. Because the initial population has very small solutions, this method incidentally finds increasingly larger expressions that lie on the non-dominated front, even though it is not explicitly minimizing the size objective. Moreover, MO GP-GOMEA continually uses fruitless search efforts to minimize the expression size further, while the smallest expression size has already been found. The individuals with the highest $R^2$ often have large tree structures that could benefit the most from optimising their expression size, but these individuals end up in the extreme cluster optimising only accuracy. We therefore conclude that for the bi-objective optimization of accuracy and expression size, an SO approach is preferable, but we also find that the clustering approach used so far has issues, which may be improved upon.

\subsection{What improvements can be derived for MO GP-GOMEA from SO GP-GOMEA?}
\label{sec:improvements}
In each of the previous experiments, the same population size (4096) was used for both the MO and SO experiments. MO GP-GOMEA however divides its population over five clusters. Hence, the extreme cluster with the accuracy objective is smaller than the total population size used in the SO approach. As population size is often a crucially important factor in what can be achieved with an EA, the population size for each EA should always be tuned individually for the fairest comparison. Therefore, we also perform an experiment in which the MO approach has a population 5 times larger (making each cluster equally large as the population in SO GP-GOMEA).

\begin{figure}[h]
  \centering
    \includegraphics[width=0.9\linewidth]{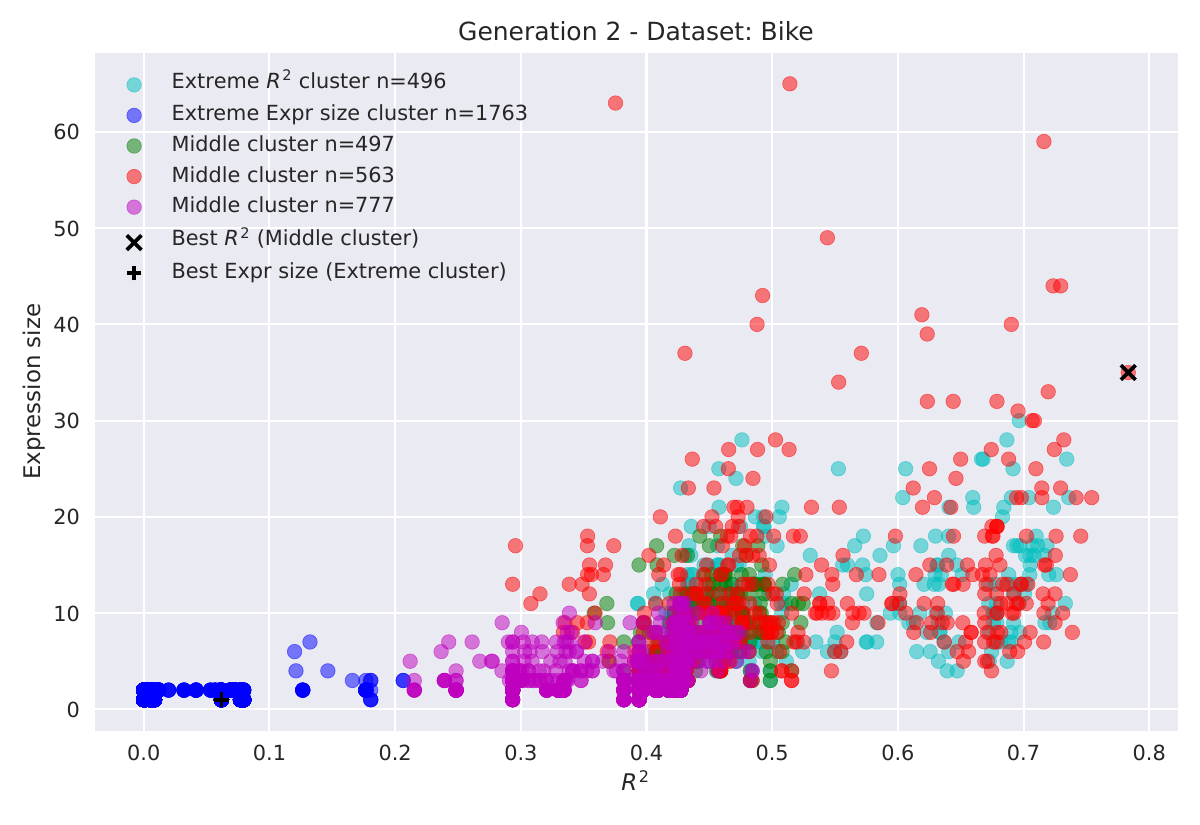}
    \vspace*{-6mm}
    \caption{Two issues in MO GP-GOMEA with the original clustering approach: the clusters are unbalanced (see the large cluster size of the expression size cluster that occurs after only 2 generations) and the solution with the best $R^2$ value (marked $\times$) is assigned to a middle cluster instead of the appropriate extreme cluster.}
    \label{fig:unbalancedfront}
\vspace*{-3mm}
\end{figure}

Cluster sizes can become skewed, as many individuals in the population end up in a cluster with other individuals that have an expression size of 1 (see Figure~\ref{fig:unbalancedfront}). This goes against the idea of spreading the search bias equally across the approximation front. To address this, we propose \textit{Balanced K-leader-means Round Robin} (BKRR) clustering and \textit{Balanced K-m-leader-means Round Robin} (BKmRR) where $m$ is the number of objectives, which in this paper is always 2, but the approach is generic for any $m$. The approaches take inspiration both from the clustering approach in the MO real-valued GOMEA~\cite{bouter_multi-objective_2017} and the original BKLM clustering method~\cite{bosman2010anticipated}.

In BKRR, we initially proceed as in the clustering approach previously used in MO GP-GOMEA. After the k-means clustering step however, we do not create donor clusters with the closest $\frac{2n}{k}$ individuals and then subsequently assign potentially unassigned individuals to random clusters and break ties randomly for individuals assigned to multiple clusters. Instead, we perform multiple assignment rounds in which we iterate over the remaining clusters, shuffling the cluster order each round. Each time a cluster is considered, we assign the closest individual in the remaining unassigned population to that cluster, in a round-robin fashion, ending up with almost exactly $\frac{n}{k}$ solutions in each cluster (deviations by 1 are possible based on population size and cluster ordering).

\begin{figure}[h]
\centering
    \includegraphics[width=0.9\linewidth]{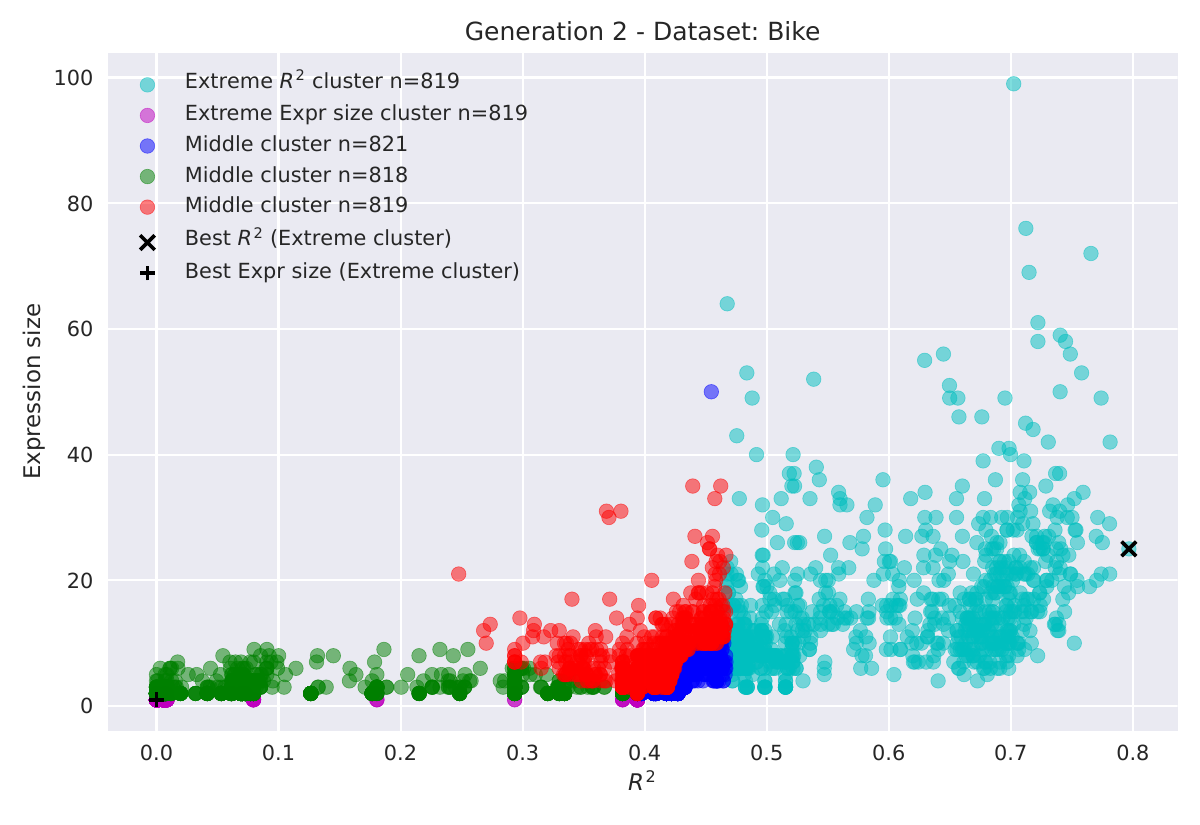}
    \vspace*{-5mm}
    \caption{Example of a balanced clustering from our proposed BKmRR clustering method. Each cluster has approximately the same number of individuals. The best solution in each objective is assigned to an appropriate SO cluster. The $R^2$ SO cluster has a visible $R^2$ cut-off around 0.48.}
    \label{fig:balancedfront}
\end{figure}

In BKmRR, we first iterate over the objectives in random order and select, for each objective, the top $\frac{n}{k}$ unassigned individuals with respect to that objective, and assign them to the extreme (SO) cluster corresponding to that objective (see Algorithm~\ref{alg:balancedkleadermeans}). The remaining unassigned individuals are then subjected to BKRR using the remaining $k-2$ clusters (i.e., excluding the individuals in the extreme clusters). The round-robin style assignment combined with the initial splitting off of extreme clusters based on their objective values for singular objectives now leads to no longer requiring donor clusters and obtaining a clustering with equal size clusters where extreme clusters are truly representative of the best solutions in that objective. For an example, see Figure~\ref{fig:balancedfront}.


\begin{table}[h]
\vspace*{-1mm}
\centering
\tabcolsep=2mm
\scalebox{0.95}{
\begin{tabular}{lcc}
\toprule
Dataset & MO with BKRR & MO with BKmRR\\
\midrule
Air          & 0.789 $\pm$ 0.016 & \textbf{0.814 $\pm$ 0.016} \\
Bike         & 0.880 $\pm$ 0.007 & \textbf{0.889 $\pm$ 0.006} \\
Concrete     & 0.841 $\pm$ 0.004 & \textbf{0.847 $\pm$ 0.009} \\
Dowchemical  & 0.809 $\pm$ 0.018 & \textbf{0.845 $\pm$ 0.007} \\
Tower        & 0.882 $\pm$ 0.008 & \textbf{0.900 $\pm$ 0.005} \\
\bottomrule
\end{tabular}
}
\caption{Comparison of the use of the two new clustering methods in MO GP-GOMEA in terms of average hypervolume ($\pm$ standard deviation). Significant results are in bold.}
\label{tab:secondary}
\vspace*{-3mm}
\end{table}

\begin{figure}[h]
\vspace*{-1mm}
\centering
    \includegraphics[width=0.85\linewidth]{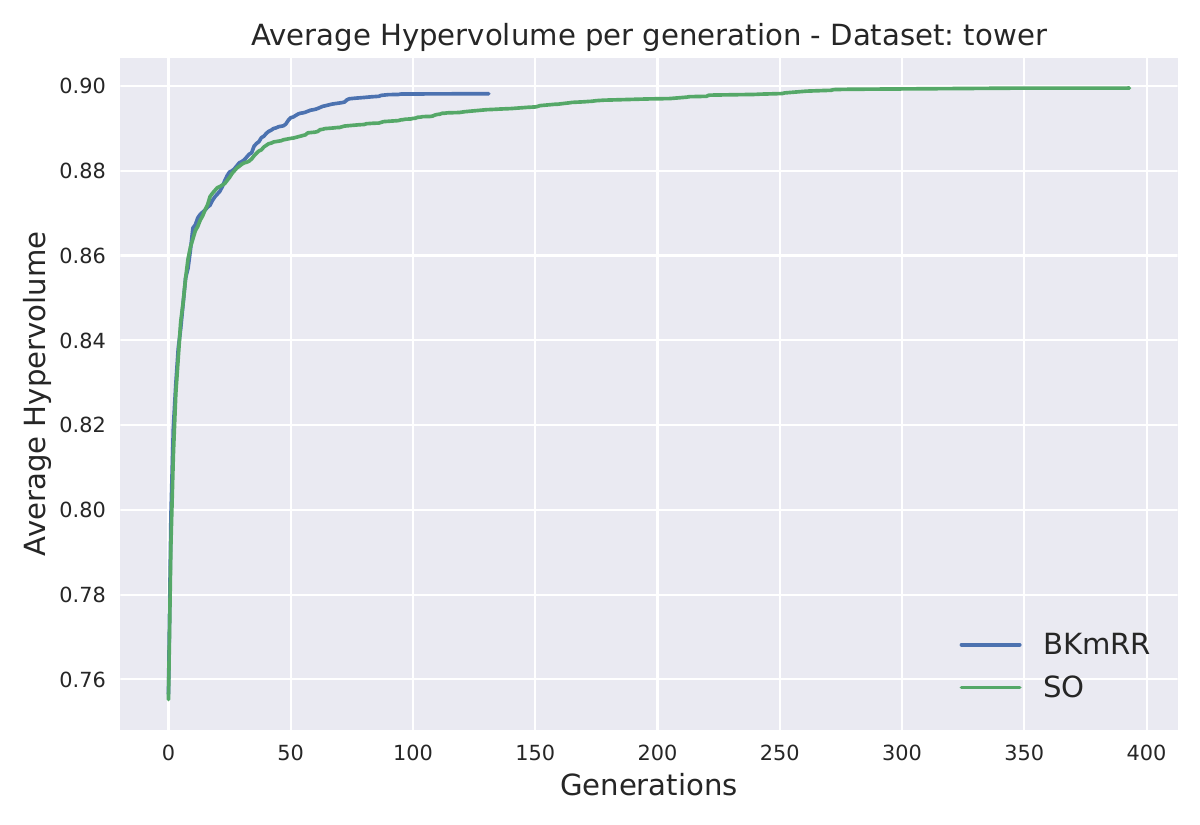}
    \vspace*{-5mm}
    \caption{Comparison of average hypervolume versus generations for MO GP-GOMEA with BKmRR and SO GP-GOMEA.}
    \label{fig:hvpergen}
\vspace*{-3mm}
\end{figure}

\begin{figure}[h]
\centering
    \includegraphics[width=0.85\linewidth]{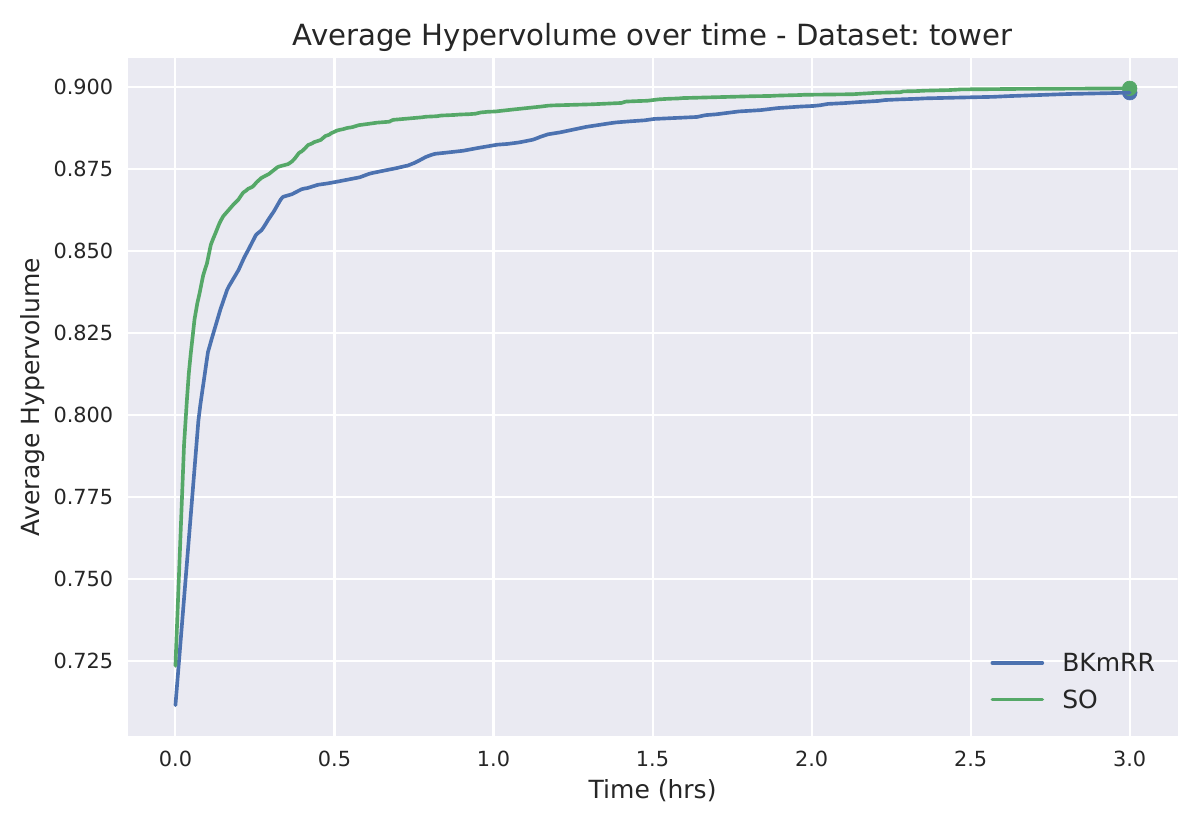}
    \vspace*{-5mm}
    \caption{Comparison of average hypervolume versus time for MO GP-GOMEA with BKmRR and SO GP-GOMEA.}
    \label{fig:hvpertime}
\end{figure}

From the results in Tables~\ref{tab:MOSO} and \ref{tab:secondary} we observe that both versions of MO GP-GOMEA that use the new clustering methods perform better than when the original clustering method is used. Moreover, the use of BKmRR statistically significantly outperforms the use of BKRR in all of the datasets.

Effectively, the BKmRR method has one extreme SO cluster that performs exactly the same task as in SO GP-GOMEA. Yet, even if the population size in MO GP-GOMEA is set so that the cluster size is the same as the population size in SO GP-GOMEA, MO GP-GOMEA is outperformed by SO GP-GOMEA in all datasets (but only significantly in the Bike and Concrete datasets). Key to understanding this difference in performance is comparing the average hypervolume measured per generation to the average hypervolume measured over time. In Figure~\ref{fig:hvpergen} we observe that per generation, MO GP-GOMEA with BKmRR closely matches SO GP-GOMEA, but, when looking at a time-based comparison in Figure~\ref{fig:hvpertime} we see that SO GP-GOMEA obtains a larger hypervolume faster. MO GP-GOMEA needs a larger population (\#clusters times larger) to construct larger expressions and also bears the overhead of constructing multiple FOSes and clustering each generation.

\subsection{Does the removal of duplicate solutions affect performance?}
In Figure~\ref{fig:balancedfront} we see that the extreme cluster with the expression size objective is balanced in the number of individuals (i.e., it has a similar number of invididuals as other clusters). However, it mostly contains duplicate expressions with a single coefficient or input feature (i.e., the size is 1, the minimum value for the size objective). These individuals will repeatedly end up in the same extreme cluster and MO GP-GOMEA will pointlessly keep trying to minimize the expression size even further. This search effort can better be spent elsewhere. One way to achieve this, is to mutate all individuals with an expression size of one, but this method does not work for other objectives where a similar problem may occur. To combat this problem in a more general fashion, each generation, we mutate all expressions that have duplicate fitness for each objective until an active node is mutated similar to work done in~\cite{miller2006redundancy}.
Symbolic expressions in GP are often modelled as binary-unary
trees.
\begin{table}[h]
\vspace*{-1mm}
\centering
\scalebox{0.95}{
\begin{tabular}{lcc}
\toprule
Dataset & MO with BKmRR mutated & SO mutated\\
\midrule
Air          & 0.819 $\pm$ 0.016 & \textbf{0.839 $\pm$ 0.008} \\
Bike         & 0.892 $\pm$ 0.005 & \textbf{0.902 $\pm$ 0.006} \\
Concrete     & 0.850 $\pm$ 0.008 & \textbf{0.876 $\pm$ 0.007} \\
Dowchemical  & 0.847 $\pm$ 0.005 & \textbf{0.863 $\pm$ 0.008} \\
Tower        & 0.897 $\pm$ 0.00 & \textbf{0.905 $\pm$ 0.006} \\
\bottomrule
\end{tabular}
}
\caption{Comparison in terms of average hypervolume ($\pm$ standard deviation) of SO GP-GOMEA and MO GP-GOMEA with BKmRR clustering, both including mutation of individuals with duplicate fitness. Significant results are in bold.}
\label{tab:duplicate}
\vspace*{-3mm}
\end{table}

Comparing the results in Tables~\ref{tab:MOSO} and~\ref{tab:duplicate}, we observe that the average hypervolume of both MO GP-GOMEA with BKmRR and SO GP-GOMEA can be improved by mutating individuals with duplicate fitness. However, SO GP-GOMEA significantly outperforms MO GP-GOMEA in all datasets.
Because there are clusters with different objectives, and individuals can switch between clusters per generation, the population in MO GP-GOMEA is less likely to converge. Conversely, SO GP-GOMEA, especially with a small population size, tends to quickly (prematurely) converge. It therefore benefits from our proposed mutations that avoid convergence.

\subsection{Can functional re-use be promoted in MO GP-GOMEA?}
While the modular version of GP-GOMEA that we use in this paper is capable of creating decomposed expressions, which may enhance interpretability of the final expression, the use of accuracy and expression size alone as objectives do not actively promote re-use of subexpressions. To address this issue, we propose a new definition of the expression size objective in which we subtract all duplicate non-leaf nodes from the total expression size, encouraging re-use. We only subtract all duplicate non-leaf nodes because leaf nodes can occur multiple times without subexpression nodes and we do not want to encourage multiple uses of leaf nodes. We refer to this new objective as the de-duplicated size objective. We use this objective only during the runtime of the algorithm, while the hypervolume is calculated using the normal size objective for comparison.

Whether the expressions underlying the real-world datasets from Table \ref{tab:realworld} contain any subexpression re-use is unknown. We therefore perform experiments in this section with five synthetic datasets from~\cite{placeholdermodularpaper} (see Appendix~\ref{sec:synthetic} in the supplementary) that are specially constructed to re-use subexpressions and have a known ground-truth. No LS terms are used or calculated in this specific set of experiments.

\begin{table}[h]
\vspace*{-1mm}
\centering
\scalebox{0.95}{
\begin{tabular}{lcc}
\toprule
Dataset & MO with BKmRR & MO with BKmRR\\
& mutated & de-duplicated mutated\\
           \midrule
Synthetic 1  & 0.962 $\pm$ 0.001 & \textbf{0.963 $\pm$ 0.001}  \\
Synthetic 2  & 0.982 $\pm$ 0.000 & 0.982 $\pm$ 0.000  \\
Synthetic 3  & 0.973 $\pm$ 0.002 & \textbf{0.975 $\pm$ 0.001}  \\
Synthetic 4  & 0.978 $\pm$ 0.000 & \textbf{0.981 $\pm$ 0.001}  \\
Synthetic 5  & 0.953 $\pm$ 0.01 & \textbf{0.974 $\pm$ 0.007}  \\
\bottomrule
\end{tabular}
}
\caption{Comparison in terms of average hypervolume ($\pm$ standard deviation) of MO GP-GOMEA with BKmRR with a normal size objective and the proposed de-duplicated size objective. Significant results are in bold.}
\label{tab:de-duplicated}
\vspace*{-5mm}
\end{table}

From Table~\ref{tab:de-duplicated} we observe that promoting re-use via the use of de-duplicated size as an objective, leads to a significantly larger average hypervolume in 4/5 synthetic datasets.

\begin{table}[h]
\vspace*{-1mm}
\centering
\scalebox{0.95}{
\begin{tabular}{lccc}

\toprule

\multirow{3}{*}{Dataset}           & \multirow{3}{*}{\begin{tabular}[c]{@{}c@{}}Avg. Diff. \\ \#subexpr. \\ used\end{tabular}} & \multirow{3}{*}{\begin{tabular}[c]{@{}c@{}}Avg. Diff. \\ \#subexpr.  \\ re-used\end{tabular}} & \multirow{3}{*}{\begin{tabular}[c]{@{}c@{}}Avg. Diff. \\ \#subexpr. \\ re-used as function\end{tabular}} \\ \\ \\

\midrule

Synthetic 1  & 0.00 & 0.40 & 0.10 \\

Synthetic 2  & -0.60 & -0.40 & -0.20 \\

Synthetic 3  & -1.40 & -1.30 & 0.00 \\

Synthetic 4  & 0.20 & -0.10 & 0.3 \\

Synthetic 5  & 0.00 & 0.00 & 0.00 \\

\bottomrule

\end{tabular}
}

\caption{The average difference in the number of (re-)used subexpressions between the normal size objective and the de-duplicated size objective (i.e., the de-duplicated count minus the normal count). A positive number indicates more reuse when using the de-duplicated size objective.}
\label{tab:subexpressions}
\vspace*{-5mm}
\end{table}

We define subexpression use as the number of times a tree representing that subexpression is called within the expanded tree representation. Re-used subexpressions are those that are invoked more than once. Functional re-use specifically refers to re-used subexpressions that are called multiple times and contain at least one non-intron argument node. For a clearer illustration, we refer to the example figures in Appendix~\ref{sec:reuseexample} in the supplementary. In Table~\ref{tab:subexpressions}, the de-duplicated size objective increases hypervolume despite leading to fewer used and re-used subexpressions in 4/5 datasets. This suggests that without it, more incorrect subexpressions are selected, lowering the achieved hypervolume.

\subsection{How does MO GP-GOMEA perform with other objectives?}
While simultaneously optimizing for both expression size and accuracy is inherently a multi-objective problem, SO GP-GOMEA with an MO archive consistently outperforms MO GP-GOMEA in terms of average hypervolume. This discrepancy can be attributed, in part, to the initialization process. GP-GOMEA intitializes using the half-and-half method. The grow part of this method can cause issues because there is a high probability of sampling a small tree. Typically, multi-objective optimization with an EA starts with a set of individuals of which the objective values are far away from the Pareto front, in all objectives. To achieve this for MO GP, this would require that for the expression size, the algorithm starts with a population of only full, large trees. However, as shown in Figure~\ref{fig:balancedfront}, within just two generations, the smallest possible expression size combined with optimal accuracy for that size, is already discovered, whereas most individuals remain far from achieving the maximum attainable accuracy for their respective size.

Mindful of the importance of the non-dominated front at initialization time, in this section we analyse various types of objectives in combination with the accuracy objective to get insights regarding whether MO GP-GOMEA or SO GP-GOMEA with an MO archive performs better if other objectives than expression size are used.

\subsubsection{Maximum error}
The average error (i.e., the standard accuracy objective that we use) and the maximum error are correlated. As the average error is minimized, the maximum error is often reduced as well. In Table~\ref{tab:maxerror}, we observe that SO GP-GOMEA still outperforms MO GP-GOMEA with this objective instead of the size objective. While SO GP-GOMEA focuses solely on optimizing the standard accuracy objective, it also minimizes the maximum error, effectively optimizing both objectives simultaneously.

\begin{table}[h]
\centering
\scalebox{0.95}{
\begin{tabular}{lcc}
\toprule
Dataset & MO with BKmRR & SO\\
& mutated & mutated\\
\midrule
Air          & 0.796 $\pm$ 0.010 & 0.817 $\pm$ 0.029 \\
Bike         & 0.888 $\pm$ 0.009 & \textbf{0.900 $\pm$ 0.008} \\
Concrete     & 0.844 $\pm$ 0.007 & \textbf{0.865 $\pm$ 0.015} \\
Dowchemical  & 0.796 $\pm$ 0.022 & 0.809 $\pm$ 0.029 \\
Tower        & 0.849 $\pm$ 0.012 & 0.849 $\pm$ 0.012 \\
\bottomrule
\end{tabular}
}
\caption{Comparison in terms of average hypervolume ($\pm$ standard deviation) between MO GP-GOMEA with BKmRR clustering and mutation and SO GP-GOMEA with mutation using maximum error as a second objective. Significant results are in bold.}
\label{tab:maxerror}
\vspace*{-5mm}
\end{table}

\subsubsection{Different complexity measure}
In~\cite{kommenda2015complexity}, a novel complexity measure is introduced which gives each operator in the tree a complexity score and sums them up to obtain a complexity measure for the entire tree. In a similar fashion, we give each operator its own score (see Appendix~\ref{section:customscores} in the supplementary for scores). In Table~\ref{tab:kommenda} we see that SO GP-GOMEA significantly outperforms MO GP-GOMEA in 5/5 datasets. The new complexity metric and the size objective are correlated in the sense that when the size objective is minimized the complexity metric is also minimized.

\begin{table}[h]
\vspace*{-1mm}
\centering
\scalebox{0.95}{
\begin{tabular}{lcc}
\toprule
Dataset & MO with BKmRR & SO\\
& mutated & mutated\\
\midrule
Air          & 0.820 $\pm$ 0.015 & \textbf{0.843 $\pm$ 0.013} \\
Bike         & 0.895 $\pm$ 0.007 & \textbf{0.905 $\pm$ 0.008} \\
Concrete     & 0.858 $\pm$ 0.007 & \textbf{0.873 $\pm$ 0.009} \\
Dowchemical  & 0.847 $\pm$ 0.011 & \textbf{0.868 $\pm$ 0.010} \\
Tower        & 0.898 $\pm$ 0.004 & \textbf{0.909 $\pm$ 0.005} \\
\bottomrule
\end{tabular}
}
\caption{Comparison between MO and SO with different complexity measure as second objective in terms of average hypervolume. Significant results are in bold typeface.}
\label{tab:kommenda}
\vspace*{-3mm}
\end{table}

\subsubsection{Regularization objective with LS}
Another means of performing regularization other than minimizing expression size, is to regularize the LS scaling ($b$) and offset ($a$) terms. The motivation for doing so is that while LS can greatly increase accuracy, it also can make an expression less interpretable because to extremely large numbers for the offset and/or scaling may be obtained. We therefore consider minimizing $log(1+ a^2 + (b-1)^2)$ as an objective instead of expression size. In Table~\ref{tab:lsregularization} we see again that SO GP-GOMEA always outperforms MO GP-GOMEA using this objective instead of the expression size objective. This time the correlation with the $R^2$ is not as strong, but SO GP-GOMEA within the allotted time budget is able to find a large variety of different LS terms and is able to outperform MO GP-GOMEA in terms of hypervolume by finding more accurate solutions.


\begin{table}[h]
\vspace*{-1mm}
\centering
\scalebox{0.95}{
\begin{tabular}{lcc}
\toprule
Dataset & MO with BKmRR & SO\\
& mutated & mutated\\
\midrule
Air          & 0.823 $\pm$ 0.011 & \textbf{0.841 $\pm$ 0.010} \\
Bike         & 0.901 $\pm$ 0.006 & \textbf{0.908 $\pm$ 0.005} \\
Concrete     & 0.861 $\pm$ 0.005 & \textbf{0.882 $\pm$ 0.009} \\
Dowchemical  & 0.856 $\pm$ 0.008 & \textbf{0.874 $\pm$ 0.011} \\
Tower        & 0.892 $\pm$ 0.004 & \textbf{0.909 $\pm$ 0.004} \\
\bottomrule
\end{tabular}
}
\caption{Comparison in terms of average hypervolume between MO GP-GOMEA and SO GP-GOMEA with the MSE of LS terms as second objective. Significant results are in bold.}
\label{tab:lsregularization}
\vspace*{-3mm}
\end{table}

\subsubsection{Number of cosine operators}
The previous subsections shows that it is hard to find objectives that do not correlate with accuracy or that result in the single-objective EA starting out close to the non-dominated front. As a final, perhaps somewhat contrived case, we consider maximizing the number of cosine operators. This objective is certainly not correlated with accuracy, nor is it likely to have good values for this objective upon initialization. In Table~\ref{tab:cosines}, we observe that for this trade-off, MO GP-GOMEA achieves a significantly higher hypervolume than SO GP-GOMEA with an MO archive. As shown in Appendix~\ref{sec:correlation} in the supplementary, the number of cosine operators is low at initialization and remains low in SO GP-GOMEA, as it is not actively optimized throughout evolution.

\begin{table}[h]
\vspace*{-1mm}
\centering
\scalebox{0.95}{
\begin{tabular}{lcc}
\toprule
Dataset & MO with BKmRR & SO\\
& mutated & mutated\\
\midrule
Air          & \textbf{0.616 $\pm$ 0.011} & 0.446 $\pm$ 0.159 \\
Bike         & \textbf{0.843 $\pm$ 0.007} & 0.325 $\pm$ 0.107 \\
Concrete     & \textbf{0.746 $\pm$ 0.018} & 0.468 $\pm$ 0.201 \\
Dowchemical  & \textbf{0.705 $\pm$ 0.011} & 0.232 $\pm$ 0.060 \\
Tower        & \textbf{0.770 $\pm$ 0.016} & 0.255 $\pm$ 0.044 \\
\bottomrule
\end{tabular}
}
\caption{Comparison in terms of average hypervolume between MO GP-GOMEA and SO GP-GOMEA with the number of cosines as second objective. Significant results are in bold.}
\label{tab:cosines}
\vspace*{-5mm}
\end{table}

\section{Discussion and Conclusions}
In this paper, we introduce and improve MO Modular GP-GOMEA by replacing the clustering method that was previously introduced for the non-modular MO GP-GOMEA. Specifically, we employ a similar strategy as in RV-GOMEA and first determine extreme clusters that excel in individual objectives before clustering the rest of the population in objective space. We find that our new clustering methods improve the performance of MO Modular GP-GOMEA. Furthermore, we tackled the issue of clusters containing individuals with duplicate fitness by introducing a mutation strategy.

While these improvements enhanced MO Modular GP-GOMEA, SO Modular GP-GOMEA with an MO archive generally outperformed it in terms of average hypervolume across most datasets within a time budget of three hours when optimizing for expression size and expression accuracy. The main reason is that SO Modular GP-GOMEA effectively finds many individuals that lie on the non-dominated front even though it focuses exclusively on optimizing accuracy. Due to the half-and-half initialization strategy used in this paper, individuals of the smallest size are already found at initialization and diverting search effort towards further minimizing the size of individuals is a waste of effort with a small time budget. The superiority of SO GP-GOMEA was also observed with the regularization of LS terms as a secondary objective, but not for the number of cosines second objective. The reason here is that for the former case, upon initialization already solutions are found with good values for the objective not being optimized on, whereas this is not so for the latter case. With a longer time budget, however, MO optimization can surpass SO in identifying solutions with comparable accuracy and reduced size. Additionally, MO performs better than SO when the secondary objective is uncorrelated with accuracy or individuals do not have objective values near the non-dominated front at initialization.

Initialising the individuals differently, e.g., uniformly in terms of size as done in~\cite{ramos2020grammatically} or~\cite{lample2019deep}, may potentially positively influence the performance of MO GP-GOMEA, but this requires further research.

Based on our findings, we recommend using SO optimization with an MO elitist archive in GP, particularly when using GP-GOMEA, as a first step for most scenarios.


\bibliographystyle{ACM-Reference-Format}
\bibliography{refs}

\clearpage

\appendix
\section{Scores of complexity measure}
\label{section:customscores}
\begin{table}[ht]
\centering
\renewcommand{\arraystretch}{1.1} 
\begin{tabular}{cc}
\toprule
\textbf{Operator} & \textbf{Complexity Score} \\ 
\midrule
Add (\(+\))       & \(c_0 + c_1\)            \\ 
Sub (\(-\))       & \(c_0 + c_1\)            \\ 
Mul (\(*\))       & \(c_0 \times c_1 + 1\)   \\ 
Div (\(/\))       & \(c_0 + c_1 + 60\)       \\ 
Neg (\(-\))       & \(c_0 + 1\)              \\ 
Sin (\(\sin\))    & \(c_0^3\)                \\ 
Cos (\(\cos\))    & \(c_0^3\)                \\ 
Exp (\(\exp\))    & \(c_0^3\)                \\ 
Log (\(\ln\))     & \(c_0^3\)                \\ 
Sqrt (\(\sqrt{}\)) & \(c_0^2\)              \\ 
Square (\(**2\))  & \(c_0^2\)                \\ 
Cube (\(**3\))    & \(c_0^3\)                \\ 
Max (\(\max\))    & \(c_0^2\)                \\ 
Min (\(\min\))    & \(c_0^2\)                \\ 
Inv (\(1/x\))     & \(c_0 \times c_1 + 1\)   \\ 
Abs (\(|x|\))     & \(c_0^2\)                \\ 
Pow (\(\text{pow}\)) & \(c_0^3\)            \\ 
Feat (\(x_i\))    & \(2\)                    \\ 
Const (\(c\))     & \(1\)                    \\ 
\bottomrule
\end{tabular}
\caption{Operators and their complexity scores. $c_0$ and $c_1$ indicate placeholders for the complexity scores of the left and right incoming child operator respectively.}
\label{tab:operator_complexity}
\end{table}
\section{Synthetic datasets}
\label{sec:synthetic}
The following expressions from~\cite{placeholdermodularpaper} were used to generate synthetic datasets with subexpression re-use and a known ground-truth. Each synthetic dataset consists of 1000 samples.

\begin{enumerate}
    \item \[
    \sum_{i=1}^8 \mathrm{sin}(x_i + x_0)
    \]
    \item \[
    \mathrm{sin}(x_2 \times x_3) + \sum_{i=1}^7 \mathrm{sin}(x_i \times x_0)
    \]
    \item \[
    \sum_{i=1}^4 \sqrt{|\mathrm{sin}(x_i \times x_0)|}
    \]
    \item \[
    f_0(f_1(x_0, x_1), f_1(x_2, x_3)) + f_1(f_0(x_0, x_1), f_0(x_2, x_3))
    \]
    where:
    \[
    f_0(a, b) = \mathrm{sin}(a + b), \quad f_1(a, b) = \mathrm{cos}(a \times b)
    \]
    \item \[
    f_0(a, b, c) = \mathrm{cos}\left(a \times \mathrm{sin}\left(\frac{b}{c}\right)\right)
    \]
    \[
    f_0(x_0, x_1, x_2) + f_0(x_0, x_2, x_1) + f_0(x_1, x_0, x_2) + f_0(x_1, x_2, x_0)
    \]
\end{enumerate}

Where \(x_i\) is the \((i + 1)\)th prime number (2, 3, 5, 7, 11, 13, 17, 19) multiplied by a random number sampled from \(\mathcal{U}[0, 1]\).
\section{Distribution of objective values}
\label{sec:correlation}
In this section, we present two examples illustrating the distribution of objective values along the non-dominated front after one generation and in the final generation, across 10 repetitions.




\begin{figure}[h]
\centering
    \begin{subfigure}{0.49\linewidth}
        \centering
        \includegraphics[width=\linewidth]{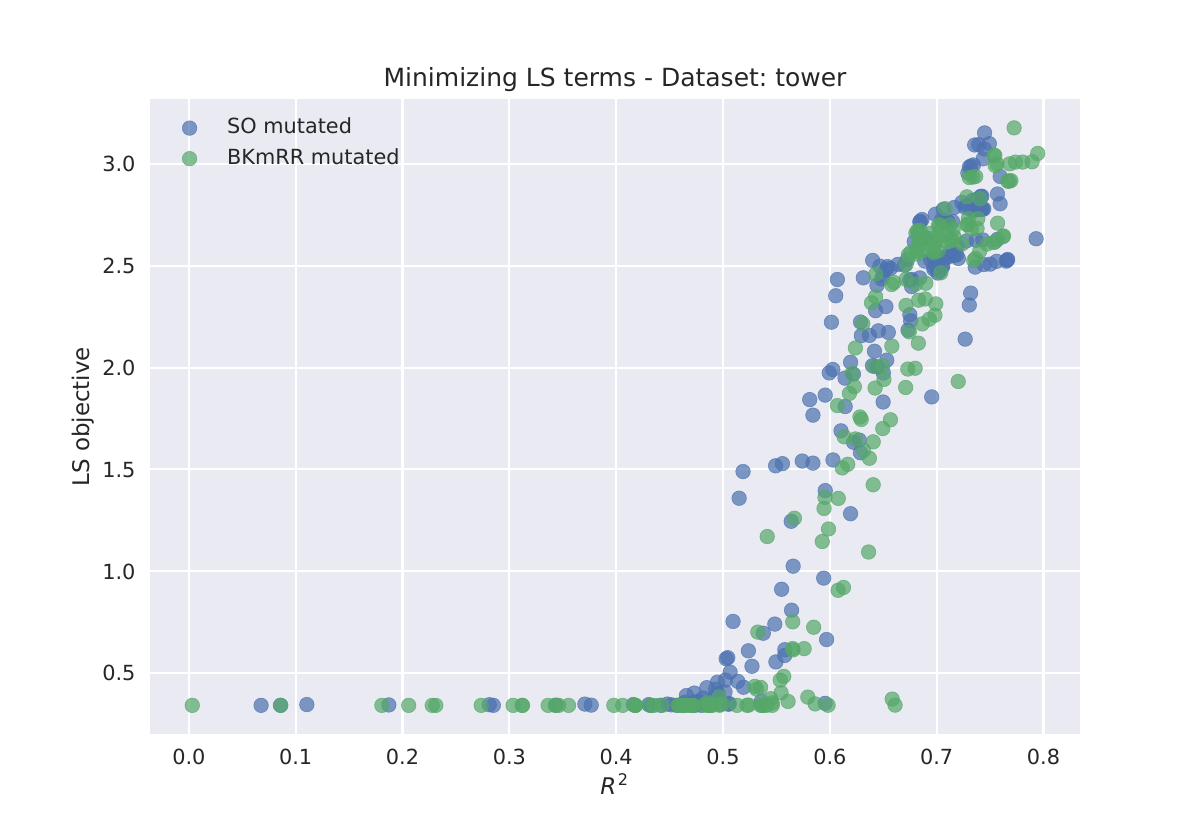}
        \caption{Start}
        \label{fig:distls0}
    \end{subfigure}
    \hfill
    \begin{subfigure}{0.49\linewidth}
        \centering
        \includegraphics[width=\linewidth]{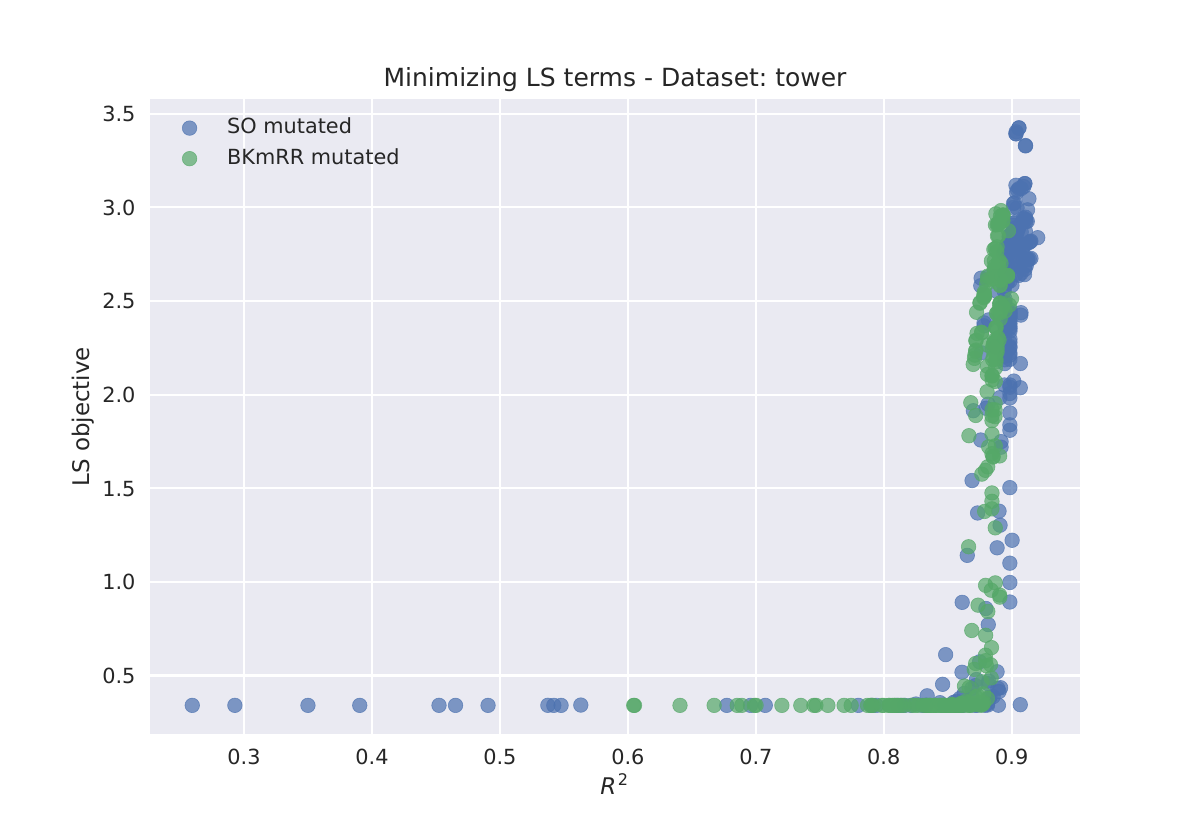}
        \caption{End}
        \label{fig:distls-1}
    \end{subfigure}
    \vspace*{-3mm}
    \caption{Scatter plot of objective values at the start (left) and at the end (right)
    of a run of Modular MO GP-GOMEA using
    regularization of LS as a second objective (10 runs combined).}
    \label{fig:distls}
\end{figure}

\begin{figure}[h]
\centering
    \begin{subfigure}{0.49\linewidth}
        \centering
        \includegraphics[width=\linewidth]{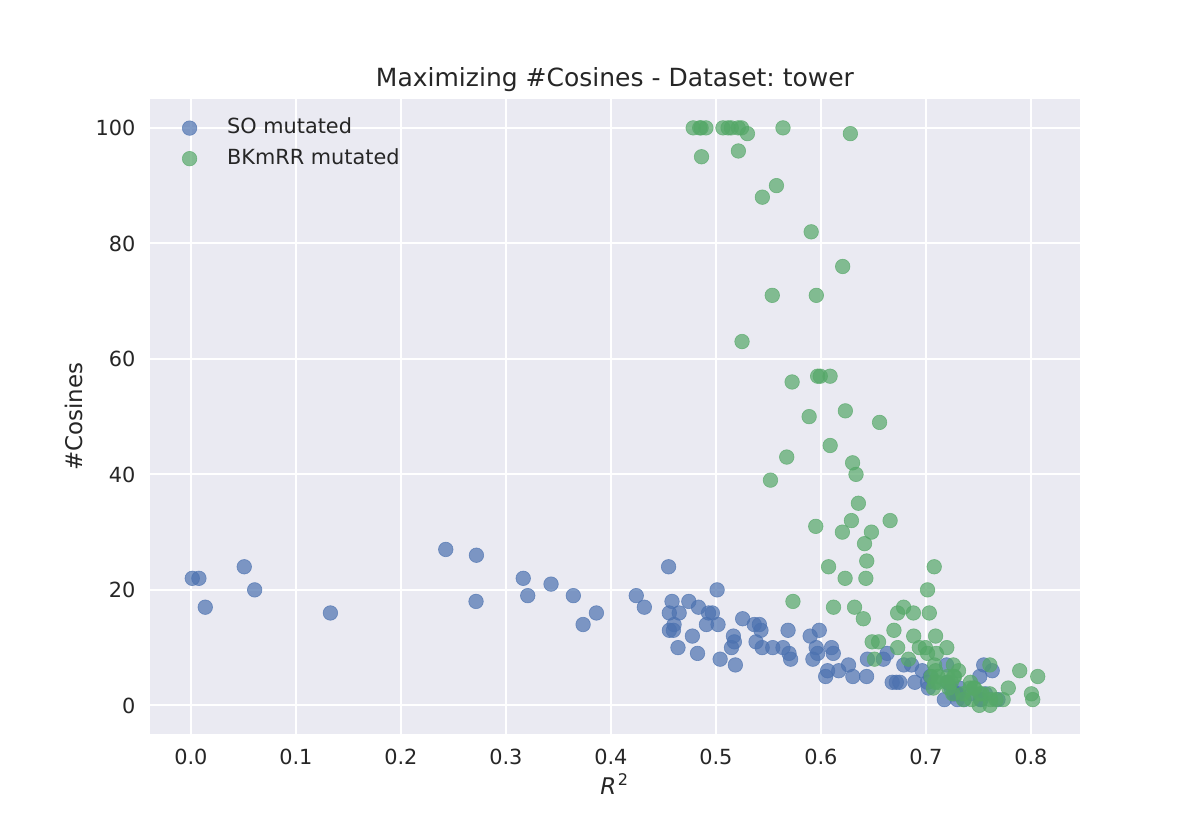}
        \caption{Start}
        \label{fig:distcos0}
    \end{subfigure}
    \hfill
    \begin{subfigure}{0.49\linewidth}
        \centering
        \includegraphics[width=\linewidth]{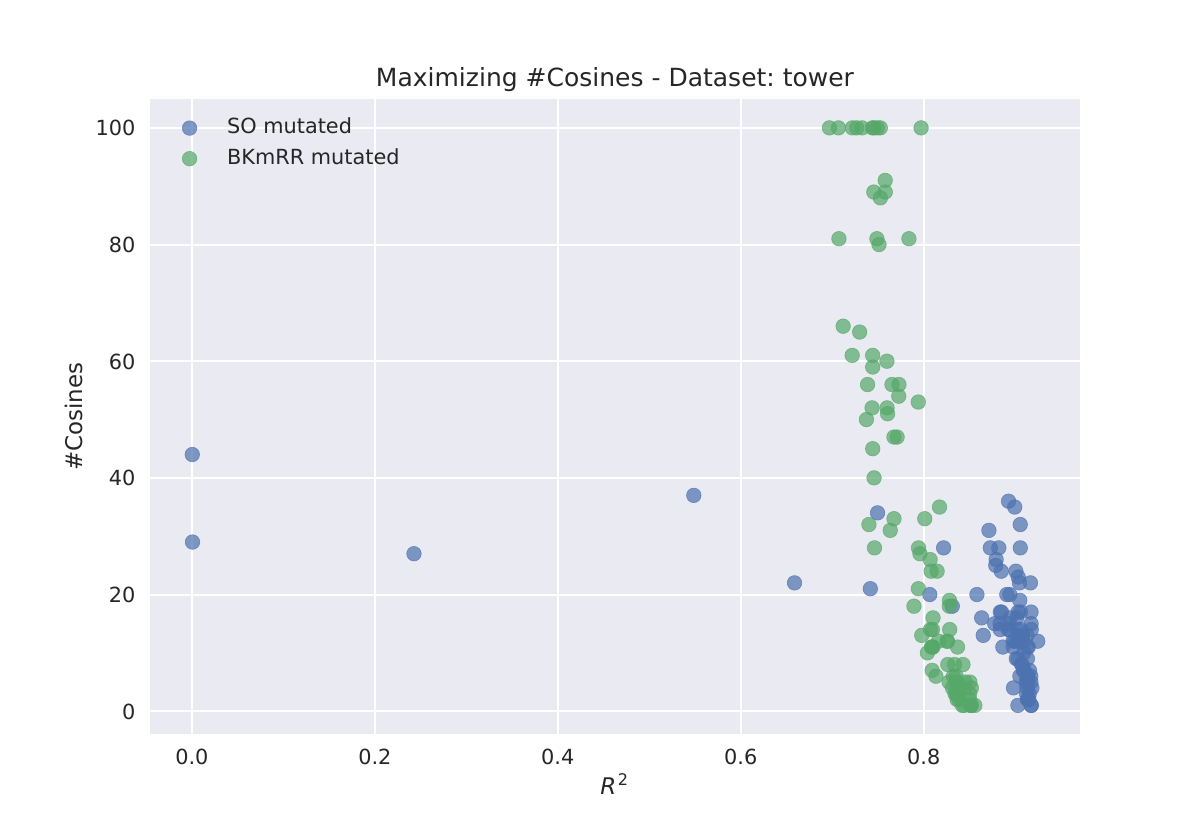}
        \caption{End}
        \label{fig:distcos-1}
    \end{subfigure}
    \vspace*{-3mm}
    \caption{Scatter plot of objective values at the start (left) and at the end (right)
    of a run of Modular MO GP-GOMEA using
    the number of cosines as a second objective (10 runs combined).}
    \label{fig:distcos}
\end{figure}



\section{Example subexpression (re-)use and functional subexpression re-use}
\label{sec:reuseexample}
\begin{figure}[h]
\centering
    \vspace*{-3mm}
    \includegraphics[width=0.9\linewidth]{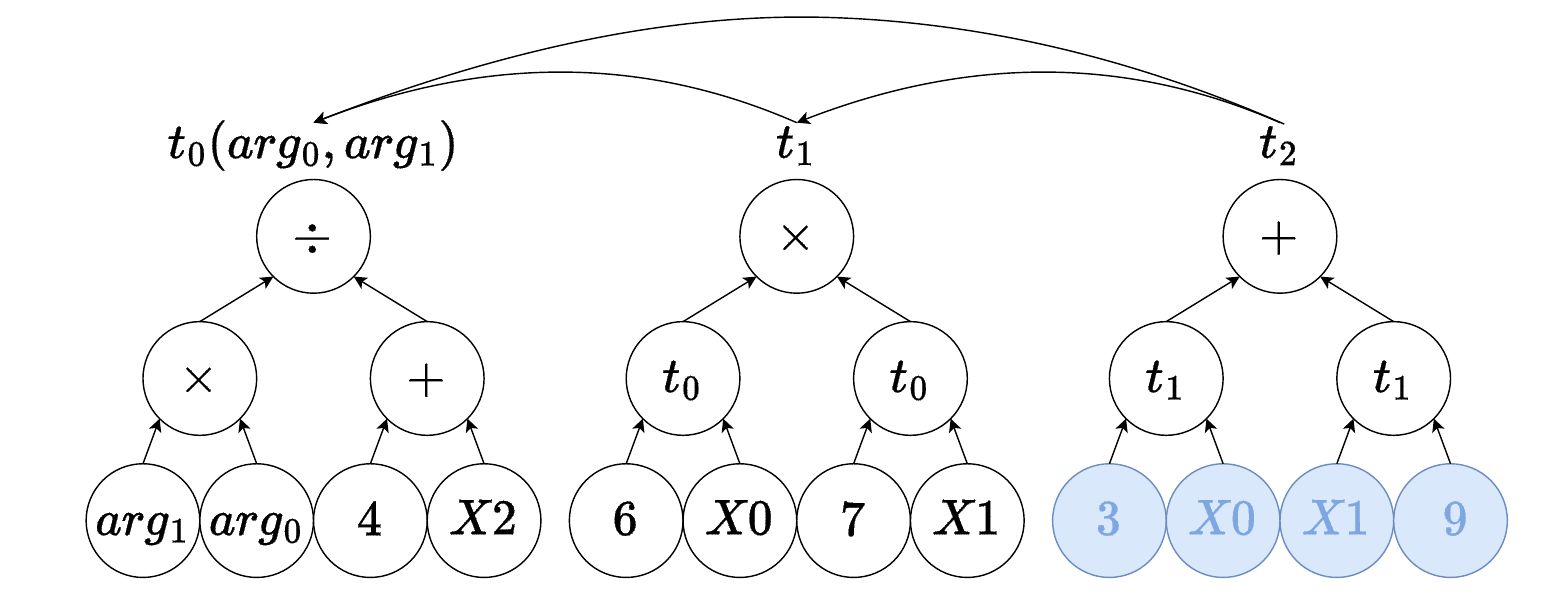}
    \vspace*{-3mm}
    \caption{Illustration of (functional) subexpression (re-)use. Note: blue nodes are introns.}
    \label{fig:examplereuse}
    \vspace*{-3mm}
\end{figure}
When we expand the main tree $t_2$ in Figure~\ref{fig:examplereuse}, we observe six instances of subexpression usage: $t_1$ twice within $t_2$, and $t_0$ appearing four times in the two uses of $t_1$ (in turn, in $t_2$). The subexpression $t_0$ is re-used three times, while $t_1$ is re-used once, resulting in a total of four re-uses, of which three are functional re-uses of $t_0$. Notably, while $t_1$ is re-used, it consistently produces the same output because its children are introns—meaning $t_1$ has no argument nodes.

\end{document}